\documentclass[sigconf]{acmart}

\usepackage{booktabs} 
\usepackage[english]{babel}
\usepackage[utf8x]{inputenc}
\usepackage[T1]{fontenc}
\usepackage{amsmath}
\usepackage{graphicx}
\graphicspath{ {files/} }
\usepackage{subfigure}

\usepackage[colorinlistoftodos]{todonotes}

\copyrightyear{2018} 
\acmYear{2018} 
\setcopyright{acmlicensed}
\acmConference[ICMR '18]{2018 International Conference on Multimedia Retrieval}{June 11-14}{Yokohama, Japan}
\acmBooktitle{ICMR '18: 2018 International Conference on Multimedia Retrieval, June 11--14, 2018, Yokohama, Japan}
\acmPrice{15.00}
\acmDOI{10.1145/3206025.3206060}
\acmISBN{978-1-4503-5046-4/18/06}

\begin{document}
\title{Automatic Prediction of Building Age from Photographs}


\author{Matthias Zeppelzauer}
\orcid{0000-0003-0413-4746}
\affiliation{%
  \institution{St. Pölten University of Applied Sciences}
}
\email{matthias.zeppelzauer@fhstp.ac.at}
\author{Miroslav Despotovic}
 \affiliation{%
   \institution{Kufstein University of Applied Sciences}
}
\email{miroslav.despotovic@fh-kufstein.ac.at}
\author{Muntaha Sakeena}
 \affiliation{%
  \institution{St. Pölten University of Applied Sciences}
 }
\email{muntaha.sakeena@fhstp.ac.at}
\author{David Koch}
 \affiliation{%
   \institution{Kufstein University of Applied Sciences}
}
\email{david.koch@fh-kufstein.ac.at}
\author{Mario Döller}
 \affiliation{%
   \institution{Kufstein University of Applied Sciences}
}
\email{mario.doeller@fh-kufstein.ac.at}

 
 





\begin{abstract}

We present a first method for the automated age estimation of buildings from unconstrained photographs. To this end, we propose a two-stage approach that firstly learns characteristic visual patterns for different building epochs at patch-level and then globally aggregates patch-level age estimates over the building. We compile evaluation datasets from different sources and perform an detailed evaluation of our approach, its  sensitivity to parameters, and the capabilities of the employed deep networks to learn characteristic visual age-related patterns. Results show that our approach is able to estimate building age at a surprisingly high level that even outperforms human evaluators and thereby sets a new performance baseline. This work represents a first step towards the automated assessment of building parameters for automated price prediction.

\end{abstract}



\maketitle

\section{Introduction}

The vision behind this research is the automated valuation of real estate objects. The valuation of building prices is a difficult task that takes into account many aspects (e.g. location, infrastructure, building size and condition etc.) and till date is mostly performed manually by domain experts. 
The most common approach for manual real estate price estimation are hedonic price models, which are regression models that take building-specific parameters 
into account \cite{liao2012hedonic}. Research on hedonic price models has shown that building age has a strong influence on real estate and rent prices \cite{zietz2008determinants,brunauer2010additive}. Age information of buildings is, however, often not available.

\begin{figure}
	\centering
    \subfigure[]{
		\includegraphics[height=0.13\textwidth]{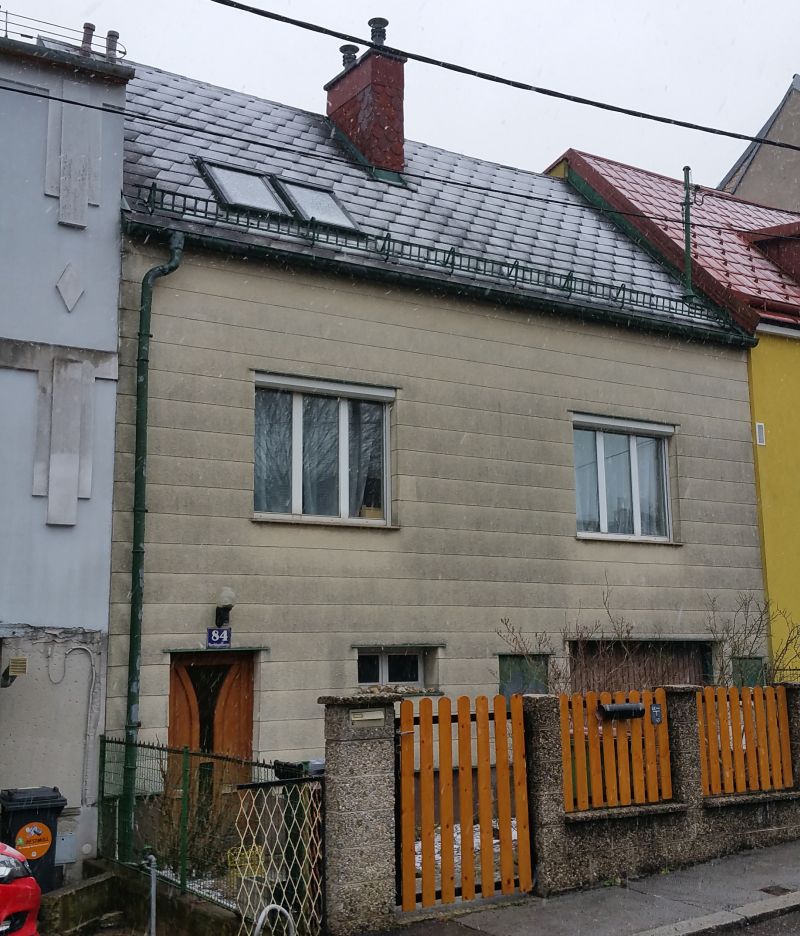}
    	\label{sfig:1960}
		}
    \subfigure[]{
		\includegraphics[height=0.13\textwidth]{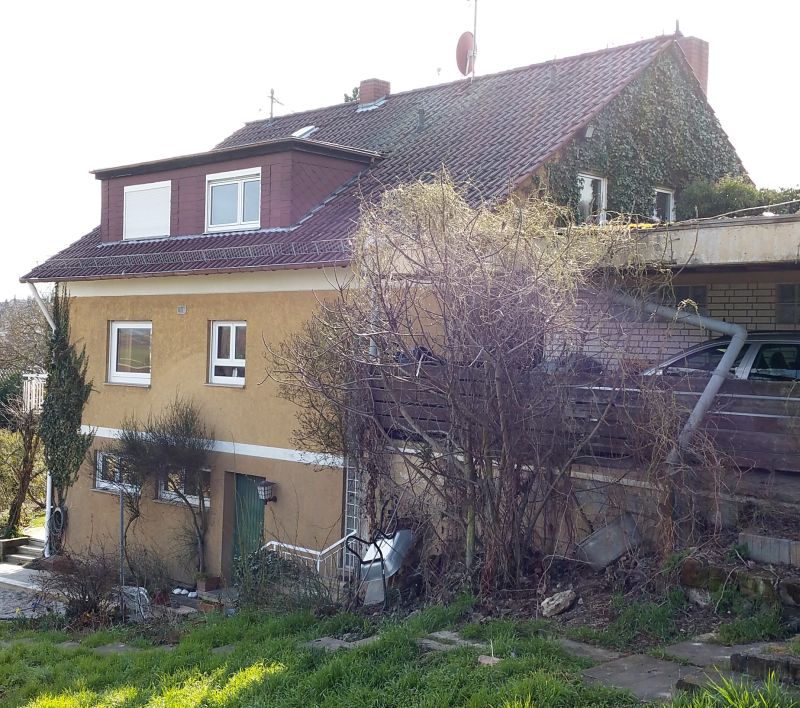}
    	\label{sfig:1974}
		}
    \subfigure[]{
		\includegraphics[height=0.13\textwidth]{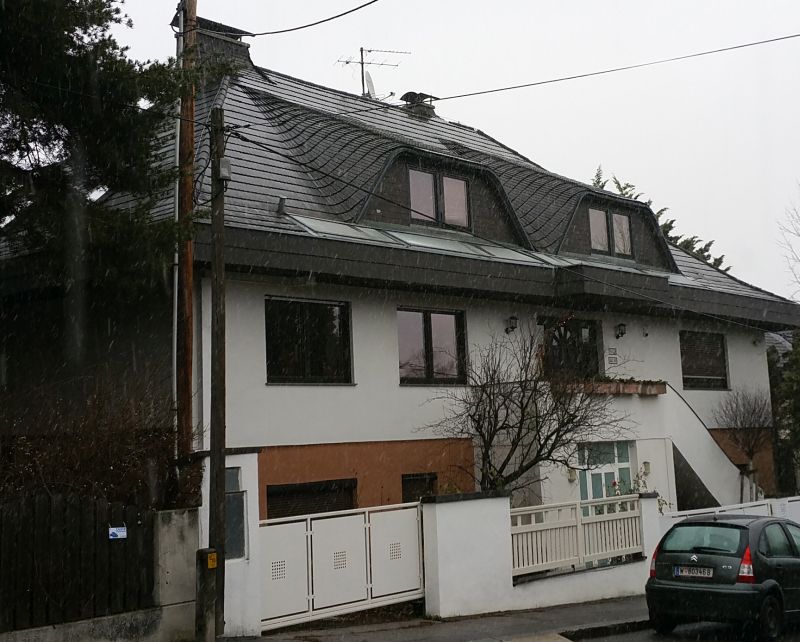}
    	\label{sfig:1984}
		}
    \subfigure[]{
		\includegraphics[height=0.11\textwidth]{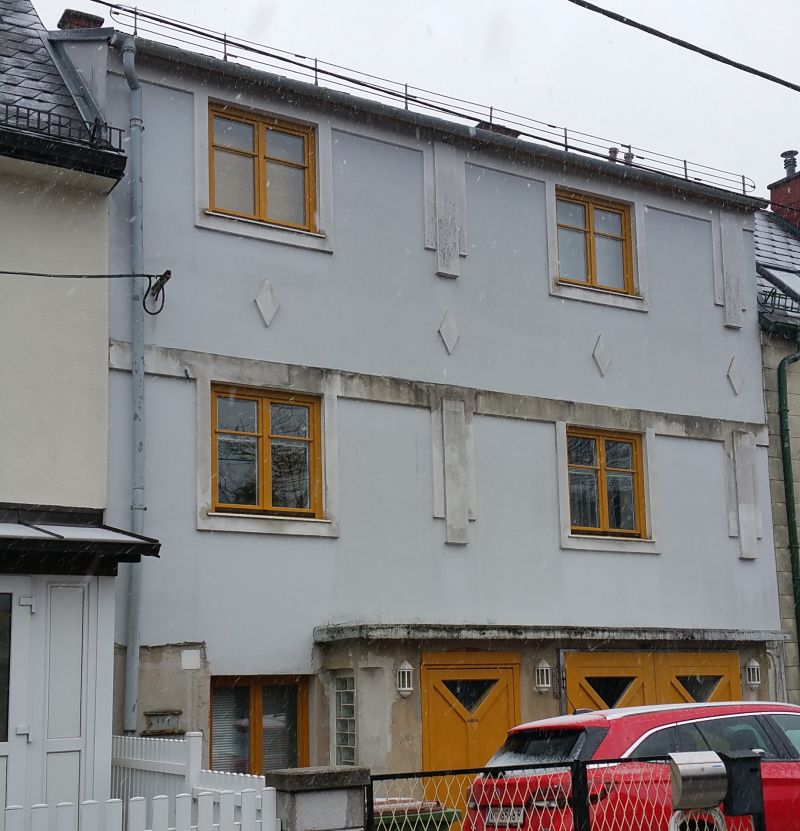}
    	\label{sfig:1993}
		}
    \subfigure[]{
		\includegraphics[height=0.11\textwidth]{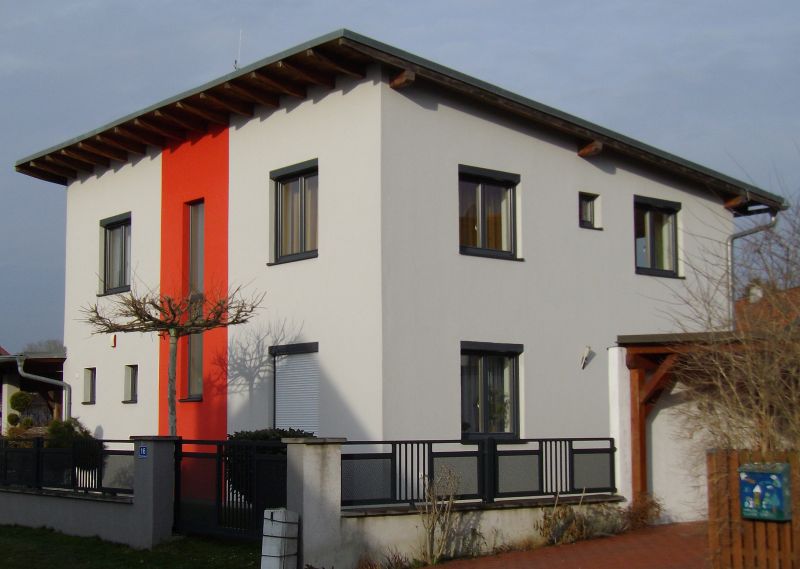}
    	\label{sfig:2007}
		}
    \subfigure[]{
		\includegraphics[height=0.11\textwidth]{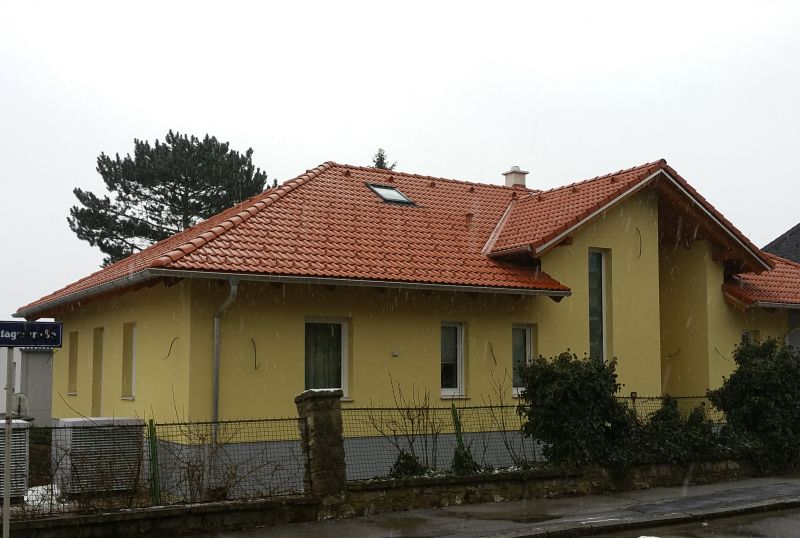}
    	\label{sfig:2014}
		}
       \caption [Caption for LOF]{Buildings from different building epochs starting with 1960 until today. Which building does stem from which epoch? The answer is in the footnote\footnotemark.}
    \label{fig:intro}
\end{figure}

The goal of this research is to improve real estate price estimation by automatically extracting the age of buildings and thereby mitigating the need for additional meta-data. We suggest that building age can be extracted by analyzing images showing the external view of houses. 
Figure \ref{fig:intro} shows an example of six different buildings from different building epochs. For a human observer, it is relatively easy to distinguish a rather old building from a newer one. The assignment of a distinct building epoch for each building is, however, a complex task. We invite the reader to guess the approximate building epoch of each building in Figure \ref{fig:intro}. The answer is here$^1$.

\footnotetext{\label{note1}Answer: (a) 1960s, (b) 1970s, (c) 1980s, (d) 1990s, (e) 2000s, (f) 2010s}

Although the task is challenging, there are certain visual clues that provide indications about coarse age or building epoch such as  
the fiber cement plates frequently used in the 1960s in Figure \ref{sfig:1960}, the rounded dorner typical for the 1980s in Figure \ref{sfig:1984}, and the typical monopitch roof of a low-energy building from the 2000s in Figure \ref{sfig:2007}. The major research question behind this work is if we can automatically capture such characteristical patterns by automated image analysis and leverage them for the automatic prediction of building age.

We present an automated method for the estimation of building age. This task has to the best of our knowledge not been investigated before. The classification of buildings from unconstrained external views, as employed in this work, is a complex task since buildings can appear in different locations of the image, from different distances and perspectives as well as under different weather and lighting conditions. Additionally, foreground objects like cars and trees often cover significant areas of the facade. We break down the problem by first trying to learn characteristic visual primitives for each building epoch. Based on age estimations of these  primitives we derive enhanced age assessments for an entire building. 

Aside from our novel approach, the contribution of this paper is fourfold: \textit{First}, we assemble several image datasets of buildings from different data sources which contain images from six different building epochs starting with 1960s until today. \textit{Second}, we perform a comprehensive analysis of the capabilities of our method and in particular investigate its generalization ability to new data and the influence of dataset bias. 
\textit{Third}, we establish a first \textit{human baseline} for building age assessment and show that our approach clearly outperforms human evaluators. \textit{Fourth}, we analyze the employed neural networks and show that building-related visual patterns are learned at several abstraction levels.

\section{Related Work}

In recent years a lot of research has been performed on the analysis of buildings, especially from satellite imagery \cite{survey}. While this works well for the detection and segmentation of land covers \cite{muhr2017} and building footprints \cite{cohen2016rapid}, satellite images cannot provide more specific visual information like the architectural style and the current state of the building which is necessary to estimate its age. More suitable for the detection of age (and potentially other parameters in future) are photos of the external view of buildings which are provided for example by real estate companies, real estate search engines, street view services, and social media platforms. Such images are, however, in most cases user-generated and unconstrained  which impedes automated analysis.   

Related methods on the analysis of building images focus rather on the retrieval of building-related photographs than on the analysis of the buildings themselves. A popular application is to find images showing the same building from different perspectives and distances. Related approaches have been introduced in \cite{relwork_1,relwork_2,relwork_3,relwork_4}. In \cite{relwork_1}, the authors use an edge-based approach based on the Hough transform to parameterize buildings which is effective in detecting buildings with strong linear edges. In \cite{relwork_2}, the authors propose a building retrieval method based on the combination of color and line features. 
An approach for the retrieval of buildings by local oriented features (using steerable filters and maximum pooling) is proposed in \cite{relwork_3}. 
A content-based image retrieval system based on consistent line clusters that further detect the approximate region of a building has been proposed in \cite{relwork_4}.   
    
Li et al.\cite{Li_2017} propose a method to predict the type of building (e.g. residential vs. non-residential building) from  Google Street View images based on Histograms of Oriented Gradients (HOG), Scale Invariant Feature Transform (SIFT) and Fisher Vectors. 
A first approach to extract higher-level building parameters has been introduced in \cite{HED2017}, where the authors try to visually derive the approximate heating energy demand of a building from building photographs.
     
Other approaches try to segment the facade out of the image as a basis for a more detailed analysis \cite{relwork_8}. The authors use a split and merge strategy to find the facade region.  Based on vertical and horizontal line scans, patches with repetitive structures are extracted and merged to find the coarse facade region. 
Beyond this, methods for the analysis and mining of different architectural styles have been introduced. 
A first approach to classify old buildings into different styles (Gothic, Baroque, and Romanesque) has been introduced by \cite{relwork_20}. The authors extract SIFT features and Bag of Visual Words (BoVW) histograms to predict different architectural styles. 
A more recent paper applies deformable parts models (DPMs) to classify building facades into 25 different architectural styles \cite{relwork_21}. The authors extract HOG pyramids and train DPMs for each style. 
A similar approach is presented in \cite{relwork_22} where the authors first decompose facade images into basic building blocks by object localization techniques and then model the spatial relationships between detected blocks (so called blocklets). Style recognition is then formulated as matching blocklets of different buildings. 
Note that compared to classical architectural styles the intra-class heterogeneity of the building epochs addressed in this work is larger because our classes do not represent individual styles but temporal epochs which may exhibit different styles. Furthermore, similar styles may appear in different epochs making the task even more challenging.

Finding typical patterns for a given class of objects (buildings) can also be considered as a pattern mining problem. Related work includes \cite{chu2012visual_relatedWork23} where the authors represent buildings by recurring spatial configurations of neighboring local features. From these configurations they build graphs and perform graph mining to discover typical configurations. A similar approach is presented in \cite{goel2012buildings_relatedWork24}. The authors first detect local features and cluster them into visual words. Next they group frequently co-occurring visual words into pairs and mine those pairs that are most discriminatory for a given class. Based on this representation the authors mine representative image patches  for different architectural styles. 
A problem of these approaches is that they do not scale well to large training sets of thousands of images and that their level of abstraction is limited. 

An alternative approach has been introduced in \cite{paris} where the authors try to discover meaningful image patches for different cities. They search for patches in street view images which are discriminative and frequently occur for certain cities. 
Lee et al. \cite{past2pres} extend this method for the clustering of buildings into coarse age groups. For this purpose, they align a cadastre map with construction dates to street view images and generate a training data set. They apply pattern discovery similarly to \cite{paris} to extract characteristic patches for different building epochs. 

\begin{figure*} [ht]
\centering
\includegraphics[width=0.8\textwidth]{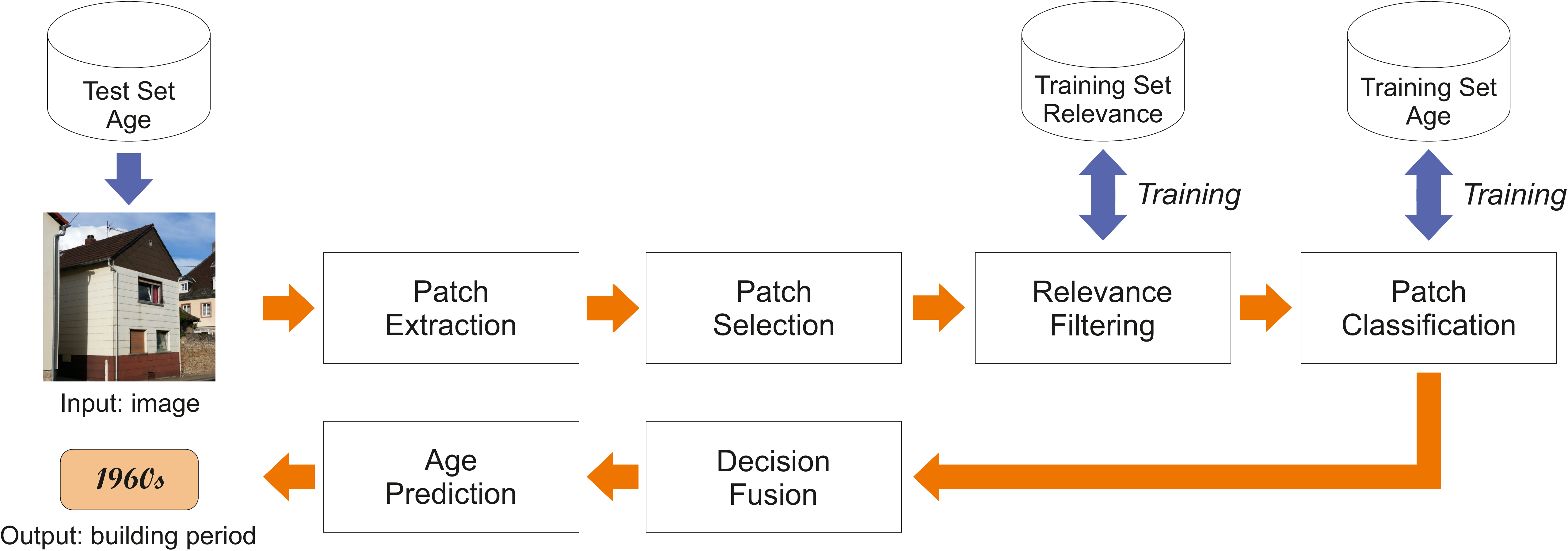}
\caption{Overview of our approach for building age estimation.}
\label{fig:approachOverview}
\end{figure*}

A limitation of the works in \cite{paris} and \cite{past2pres} is that the patch matching is based on HOG descriptors. Initial experiments with our data has shown that HOG-based patch matching yields unintuitive similarity estimates and is computationally expensive. Inspired by \cite{past2pres}, our approach introduces a number of extensions: (i) patch selection is not random but steered by the underlying structure in the images, (ii) it circumvents the limitations introduced by pair-wise HOG matching by training a classifier directly from the patches and (iii) we employ a more powerful representation of the patches based on Convolutional Neural Networks (CNNs).

\section{Approach}

The basic assumption behind our approach is that certain visual building-specific elements provide useful indicators for different building epochs and the approximate construction year. We expect the entire input image of a building 
to be too unspecific to extract age-related patterns from buildings (or rather it would need a tremendous amount of input images and training time to learn important structures completely autonomously). 
To accelerate learning and to decrease the amount of necessary training data we base the analysis on image patches. Image patches contain less complex content and may better match individual characteristic building-specific elements. Two major questions in this context are: (i) where in the image patches should best be extracted and (ii) which patches are discriminative and characteristical for a given building epoch? To answer the first question we investigate different patch sampling strategies (see Sections \ref{subsec:patchExtraction} and \ref{subsec:patchSel}). To account for the second question, we leverage Convolutional Neural Networks (CNNs) and learn patterns from the patches in a supervised manner. An overview of our method is depicted in Figure \ref{fig:approachOverview}. The input is an unconstrained image of a building from which we extract patches and heuristically select the most representative ones. Additionally, we apply a specifically trained relevance filter to remove patches not related to buildings. From the remaining patches we train a classifier for different building epochs.
Finally, decision fusion and age prediction infer the most likely building epoch for entire buildings. 

\subsection{Patch Extraction}
\label{subsec:patchExtraction}

\begin{figure}[ht]
\includegraphics[width=0.35\textwidth]{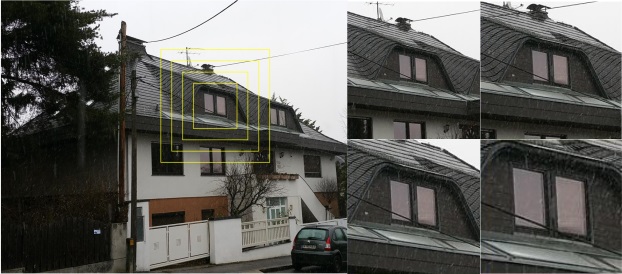}
\caption{Patches of different scales extracted at a given location (left, yellow squares), resulting rescaled patches (right).}
\label{fig:patchExtraction}
\end{figure}

In contrast to previous work on patch-based pattern mining \cite{paris,past2pres} we do not sample patches randomly from the input image because this may lead to a bad spatial coverage. First experiments trying to leverage SIFT feature detection to identify suitable patch locations led to strongly imbalanced sampling across the image and were thus discarded \cite{lowe1999object}. To obtain a better coverage and to save computation time, we apply sampling at regular intervals (as in dense SIFT) and extract overlapping patches of different scales at each sampling location, see Figure \ref{fig:patchExtraction}. This leads to a large number of candidate patches which cover the entire image at different scales. After extraction all patches are rescaled to the same size.

\subsection{Patch Selection}
\label{subsec:patchSel}

Patch selection aims at removing patches which do not represent building-related information, for example sky in the background or occluding objects like cars in the foreground of the building. This reduction of data is necessary to speed up the training of the subsequent CNNs. Furthermore, it helps to direct the further analysis to building-related content, which facilitates CNN training and enables to use smaller training sets.  
We propose two heuristics for the extraction of meaningful building-related patches.

\begin{figure}[ht]
\includegraphics[width=0.65\linewidth]{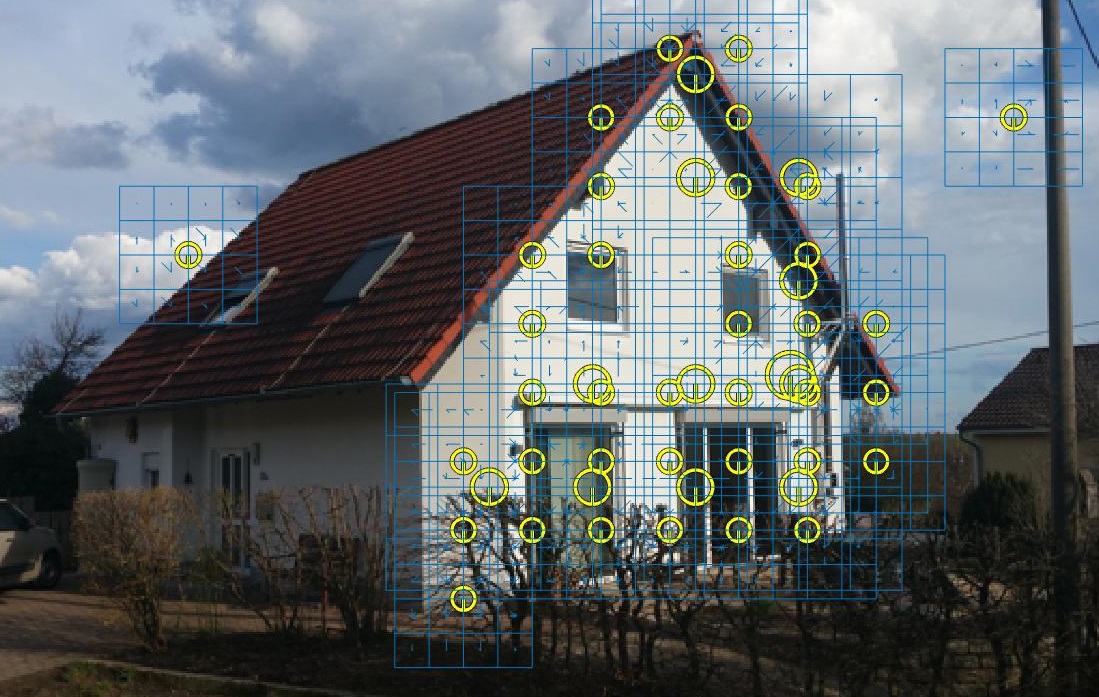}
\caption{The top 3\% patches with highest contrast according to the norm of their SIFT descriptors. The patches cover well the area of the building (facade). 
}
\label{fig:highContrastPatches}
\end{figure}

\paragraph{High-contrast patches:} patches with low contrast are likely to represent unimportant information (e.g. homogeneous sky in the background or homogeneous areas of a smooth facade). To obtain a measure for the amount of contrast inside a patch we compute the norm of its (unnormalized) SIFT descriptor, i.e. the norm of the gradient histogram. The norm provides an estimate of the amount of gradient variation inside the patch in different directions. From all input patches we select only the $t\%$ patches with the maximum norm to obtain a smaller set of high-contrast patches, see Figure \ref{fig:highContrastPatches}.

\begin{figure}[ht]
 	\centering
		\includegraphics[width=0.65\linewidth]{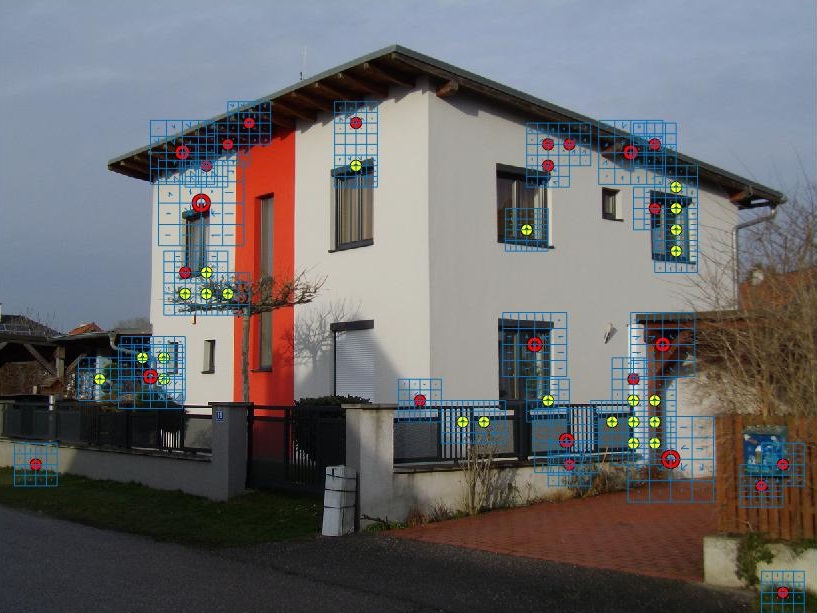}
    \caption{Different patch selection strategies: patches selected through clustering (red) represent more heterogeneous structures while high-contrast patches (yellow) exhibit higher redundancy (more repetative visual structures).}
    \label{fig:selectionStrategies}
\end{figure}

\paragraph{High-contrast patch clusters:} The first strategy may be biased towards similar and frequently occurring structures (e.g. corners of windows) resulting in visually similar image patches. To obtain more visual heterogeneity we first cluster the extracted image patches. Therefore, each patch is represented by its (normalized) SIFT descriptor. $K$-Means is applied to obtain $K$ clusters of patches for each input image. From each cluster we select the patch which is closest to its cluster center. These patches are more likely to represent visually different content. From the resulting $K$ patches, we select the  $t\%$  patches with highest contrast as in the first selection strategy. Figure \ref{fig:selectionStrategies} shows the difference between both strategies.

\subsection{Relevance Filtering}
\label{subsec:relFiltering}

A problem that remains after patch selection is that many patches represent information from the image which indeed have a lot of structure but are not related to the depicted building, e.g. parts of cars, trees, fences, etc. To better distinguish building-related from non-related patches, we train a classifier at patch-level that categorizes patches into different object types. We manually compile a training set of image patches from our training set (see Section \ref{subsec:dataset}) for 12 different object categories which often appear in outdoor photographs of buildings: cars, people, trees, grass, asphalt, poles, sunblind, furniture, fence, firewood, sign, a class containing miscellaneous patches of other type, and one positive class (building). The total training set for relevance filtering contains 8.000 
patches. As a classifier we employ a pre-trained CNN (AlexNet trained on ImageNet \cite{krizhevsky2012imagenet}) and adapt its output layer to 13 classes. We chose a multi-class setup instead of a binary setup (building vs. non-building) to enable the network to better model the structure of the data and to learn a more powerful representation. We fine-tune all layers of the network. The trained net is then applied as filter to remove all patches from further processing that are not classified as building.


\subsection{Patch Classification \& Decision Fusion}
\label{subsec:patchClass}
Each patch that remains after relevance filtering is input to patch classification. To increase heterogeneity during training we apply data augmentation. Each patch is flipped, cropped, and scaled with a random probability. Additionally, brightness, contrast, and saturation are shifted randomly to account for illumination variations.

For patch classification we employ two different CNN architectures: Alexnet \cite{krizhevsky2012imagenet} with 8 layers and ResNet \cite{He_2016_CVPR} with 50 layers. Alexnet is a widely used CNN consisting of five convolutional layers (with max pooling, local response normalization, and rectified linear units (ReLu) as activation function), three fully connected layers and a softmax layer on top. ResNet50 consists of larger stacks of convolutional layers with batch normalization \cite{ioffe2015batch}.  ResNet employs direct connections that skip individual layers (skip connections). These connections accelerate training and enable the construction of deeper networks. Although the ResNet architecture has many more layers, it has less parameters than Alexnet and is thus faster to train. 
For both employed architectures we rescale the last layer to the number of building epochs that should be distinguished.

The patches are fed into the networks with all RGB color channels and are rescaled to the input size of the networks (224x224). The networks are applied either individually (for comparison) or in combination as an ensemble. In case of ensemble classification, we fuse the outputs of the networks by averaging their respective class likelihoods from the softmax layers. The result is again a probability distribution from which we retrieve the class with the maximum likelihood as the final prediction.

\subsection{Age Prediction}
\label{subsec:agePrediction}
All analyses and predictions so far have been made at patch-level. To get a prediction for an entire building, the patch-level predictions need to be aggregated into a global estimate. We investigate two different strategies:
\begin{itemize}
\item \textit{Majority voting:} To get a robust estimate we take all patch-level predictions for an entire image and return the building epoch which is predicted most often.
\item \textit{Overall likelihood:} To get a more sensitive decision, we average the class likelihoods and return the building epoch with the highest overall likelihood for all patches.
\end{itemize}

A source of confusion may be patches with \textit{ambiguous age predictions}, i.e. patches where the most likely class $c_{max}$ is not remarkably more likely than the other classes. Such patches may represent age-invariant visual patterns that impede age prediction. We define a patch as ambiguous if the difference between the likelihood $c_{max}$ and the likelihood for the second most likely class is below a threshold $t_u$. We remove such ambiguous patches during age prediction.

\section{Experimental Setup}

\subsection{Datasets}
\label{subsec:dataset}

The employed data comes from two different sources: a database of real estate evaluation reports and from web platforms\footnote{The images used in this work are not under creative commons license and thus cannot be shared, we will, however, publicly share extracted features and trained networks under \url{https://phaidra.fhstp.ac.at/detail_object/o:2960}.}. We restrict the analysis to those building epochs for which a sufficient number of houses could be retrieved (see below).

\paragraph{Building epochs} Epochs employed in real estate industry usually capture entire decades (e.g. 1960s, 1970s...) as defined by \cite{statistikaustria}.
In contrast to the official counting where each epoch begins with the first odd year of the decade (e.g. 1961) we set the begin of each epoch to the beginning of the decade, i.e. 1960. This is due to the fact that for houses where only the decade but not the exact year of construction (YoC) is known, real estate agents and valuers often denote the YoC simply with the belonging decade. Thus, the official counting in such cases would lead to artificial class confusions.

We have compiled three datasets which all contain images of houses from all six building epochs. Due to the different availability of buildings for the different epochs, the datasets are not completely balanced, see Table \ref{tab:datasets}. 

\begin{table}[ht]
\centering

\resizebox{\linewidth}{!}{%
    \begin{tabular}{l | r r r r r r | r}
\textbf{dataset $\backslash$ epoch}  &	\textbf{1960s}	&	\textbf{1970s}	&	\textbf{1980s}	&	\textbf{1990s}	&	\textbf{2000s} 	&	\textbf{2010s} & \textbf{sum}	\\ 
\hline
training $T_{tr}$ &	649	&	1.075	&	1.378	&	1.424	&	1.315	&	579 & 6.450	\\ 
validation $T_{v}$ &	84	&	153	&	205	&	207	&	201	&	80	& 930\\ 
test $T_{t}$ &	194	&	299	&	420	&	431	&	372	&	178	& 1.894\\ \hline
web images $W_{t}$ &	209	&	269	&	198	&	229	&	227	&	253	& 1.385\\ \hline
human baseline $H_{t}$ &	47	&	53	&	61	&	118	&	139	&	52	& 470\\ 

    \end{tabular}
    }
    \caption{Datasets used for our experiments with the number of images per building epoch.}
\label{tab:datasets}
\end{table}

\vspace{-30pt}

\paragraph{Real estate valuation reports:} From a real estate valuation company we obtained a dataset of 17.241 valuation reports of single family houses generated by real estate experts. We first remove all reports that do not contain a photo of the building. The remaining reports contain at least one or more pictures, postal code, YoC, and the so called fictitious year of construction (FYoC). The FYoC is a number determined by the real estate valuers and is computed as $FYoC = YoC + N$, where N represents the number of years the total lifespan of a building has been extended by renovations and modernizations. 
Modernizations (e.g. of the facade) often change the appearance of a building significantly, making its real age difficult to discover visually. Since our method operates purely on visual information, we exclude such renovated houses from this dataset to avoid a bias during training. 
The final dataset contains 9.250 RGB images of the single family houses. For the training of our classifiers we split the dataset into three subsets: training $T_{tr}$ with 6.450 images ($\approx$70\%), validation $T_{v}$ with 930 images ($\approx$10\%) and test $T_{t}$ with 1.894 images ($\approx$20\%). Since the original dataset may contain several images for each house, we have taken special caution that no house appears in more than one partition. 

\paragraph{Web images:}

In order to evaluate the influence of the dataset bias \cite{torralba2011unbiased} to our approach, we have compiled a separate dataset by crawling images together with meta-data from real estate offers on respective websites. Only offers where an exact YoC was available were crawled. Redundant offers (from the same or several platforms) were removed. Overall, a set of 1.385 images has been compiled, which we refer to as $W_{t}$. This dataset serves as an additional independent test set. 
Note that in this dataset we cannot guarantee that $YoC$ equals $FYoC$. Thus, renovated buildings with a changed (and thus unauthentic) visual appearance may be included. This strongly increases the complexity of the dataset. %

\paragraph{Human Baseline Data:}
\label{subsec:humanBaselineSetup}
To get a better feeling for the complexity of the task of building age estimation, we establish a human baseline for building age prediction. 
To this end, we assemble a third dataset which is again disjoint to all other sets. The dataset stems from additional (previously not used) 
real estate reports and comprises of images from 470 distinct houses from all six epochs and is referred to as $H_t$ in the following. To establish a fair and objective comparison with our approach we take our pre-trained method and apply it directly on this dataset.

\subsection{Method Setup \& Training}
\label{subsec:methodSetup}

The proposed method has a number of parameters that are defined as follows. For patch extraction, we employ square patches of side length 16, 24, 32, and 40 pixels (see Figure \ref{fig:patchExtraction}) and sample them  for each scale continuously over the image with 50\% horizontal and vertical overlap. SIFT descriptors are extracted with standard parameters \cite{vedaldi08vlfeat}. For the first patch selection strategy (high-contrast patches) we employ $t=0.5\%$, $t=1\%$, and $t=1.5\%$ of high-contrast patches to evaluate the sensitivity of parameter $t$. For the second selection strategy (high-contrast patch clusters) we employ $K=50$ clusters for each image and percentages of 7\%, 14\% and 21\% for $t$.

For \textit{relevance filtering} we employ Alexnet \cite{krizhevsky2012imagenet} pre-trained on ImageNet adapted to 13 selected object classes (see Section \ref{subsec:relFiltering}). We train the network for 20 epochs with a learning rate of 0.0001, decay of 0.0005, momentum of 0.9 and a batch size of 256. 

For \textit{patch classification} we employ extensive data augmentation (see Section \ref{subsec:patchClass}) during training for both employed networks Alexnet and ResNet (both pre-trained on ImageNet). For both nets the output is reduced to 6 neurons representing the six building epochs. Common training parameters for both nets are 20 epochs with a learning rate of 0.0001, decay of 0.0005, momentum of 0.9 and a batch size of 256. 
All nets were trained with cross-entropy loss and stochastic gradient descent \cite{bottou2010large}. 

For the detection of ambiguous patches, we set parameter $t_u=0.25$ (empirically estimated on validation set).
All experiments were performed in MATLAB on a workstation with Ubuntu 14 OS, 64 GB RAM and an NVIDIA T1080. The MatConvNet framework was used for training the networks \cite{vedaldi2015matconvnet}.

\subsection{Evaluation \& Research Questions}
\label{subsec:RQs}

The primary goal of the evaluation is to investigate how well the age of a building can be estimated from a photo. To assess the performance we train our method on training set $T_{tr}$ and employ the validation set $T_v$ to monitor training progress and overfitting behavior. Finally, we evaluate the method with the independent test set $T_t$. We employ the classification accuracy as well as top-1 error (=1-accuracy) as performance measures for all experiments at patch-level as well as on image-level. Accuracy represents the portion of correctly classified images in the entire test set. Note that, classification accuracy is a reasonable choice since the dataset is mostly balanced.

Beyond pure classification performance, we aim at answering the following research questions (RQ):
\begin{itemize}
  
  \item \textbf{RQ1}: How large is the generalization ability and is there a notable dataset bias, i.e., do results generalize to data from different sources?
  \item \textbf{RQ2}: How well does the method perform compared to human evaluators?
  \item \textbf{RQ3}: Can we automatically discover characteristic age-related visual primitives from unconstrained building photographs
\end{itemize}

To account for \textbf{RQ1} we evaluate the approach on the independent partition $T_t$ (test set) and on the data from the second platform $W_{t}$ to investigate dataset bias (see Section \ref{subec:genealization}). \textbf{RQ2} is investigated by letting expert users guess building epochs on the human baseline $H_t$, see Section \ref{subsec:HumanBaselineresults}. To answer \textbf{RQ3}, we visually inspect those patches from the test set $T_t$ which obtain the highest likelihoods for a building epoch. Additionally, we investigate neural network activations and visualize the learned filters (see Section \ref{sec:qualitativeResults}).

\section{Results}

\subsection{Parameter Sensitivity}
\label{sec:preliminary}
In a preliminary evaluation, we assess the sensitivity of our method to its parameters. Especially the parameters for patch selection and extraction are essential, as they strongly influence the selection of training patches for the subsequently trained networks and thereby the overall performance. Table \ref{tab:sensitivity} summarizes the results of the experiments and shows that larger percentages, i.e. larger numbers of patches lead to slightly better (lower) top-1 error for the validation set. Furthermore, the selection strategy with patch clustering provides better results than just selecting the highest contrast patches globally. This is most likely due to the better visual heterogeneity obtained by clustering. Finally, relevance filtering ``RF'' improves results consistently by a small margin.

\begin{table}[ht]
\centering
    \begin{tabular}{l | r | c | r | c | c }

    Patch selection		& Perc. 	& RF. 		& \#patches & $err_{trn}$ 	& $err_{val}$ \\ \hline
    Hi-C. 			& 0.5\% 	& N 		& 86.596 	& 	0.49			&	0.65		\\ 
    Hi-C. 			& 1\% 		& N 		& 181.676 	& 	0.38			&	0.61		\\ 
    Hi-C. 			& 1.5\% 	& N 		& 267.552 	& 	0.31			&	0.62		\\ \hline
    Hi-C. Cluster 	& 7\% 		& N 		& 102.328 	& 	0.53			&	0.63		\\ 
    Hi-C. Cluster 	& 14\% 		& N 		& 179.208 	& 	0.48			&	0.61		\\ 
    Hi-C. Cluster 	& 21\% 		& N 		& 274.585 	& 	0.43			&	0.60		\\ \hline	
    Hi-C. 			& 0.5\% 	& Y 		& 34.842 	& 	0.52			&	0.61		\\ 
    Hi-C. 			& 1\% 		& Y 		& 73.779 	& 	0.44			&	0.60		\\ 
    Hi-C. 			& 1.5\% 	& Y 		& 109.714 	& 	0.39			&	0.60		\\ \hline	
    Hi-C. Cluster 	& 7\% 		& Y 		& 48.735 	& 	0.54			&	0.63		\\ 
    Hi-C. Cluster 	& 14\% 		& Y 		& 86.208 	& 	0.50			&	0.59		\\ 
    Hi-C. Cluster 	& 21\% 		& Y 		& 137.808 	& 	0.44			&	0.58		\\ 
    \end{tabular}
    \caption{Training results for different patch selection strategies (high-contrast patches, ``Hi-C.'' and high-contrast patch clusters ``Hi-C. Cluster'' with K=50), different percentages ``Perc.'', as well as with and without relevance filtering ``RF''.  ResNet with 10 epochs is used as network model. We provide the number of resulting patches, the top-1 error $err_{trn}$ for the training set $T_{tr}$ and the top-1 error $err_{val}$ for the validation set $T_v$.}
	\label{tab:sensitivity}
\end{table}

Experiments presented in Table \ref{tab:sensitivity} were obtained after 10 epochs of fine-tuning ResNet. We take the best result obtained so far (last row in Table \ref{tab:sensitivity} and continue the training for additional 30 epochs to see if performance further improves. Retraining mainly reduces the training error (down to 14\%). The validation error decreases only slightly and starts to oscilate around 58\%. This shows that the network starts to overfit on the training data after 10 epochs. 

From Table \ref{tab:sensitivity} we observe that the validation errors are relatively high compared to the training errors (around 60\%, corresponding to a classification accuracy of 40\%). Note that both measures are computed at patch-level and are thus not directly representative for the overall task of predicting the age of an entire building. 

\subsection{Age Prediction}
\label{subsec:agePredictionResults}

\begin{figure}[ht]
 	\centering
 		\includegraphics[width=0.4\textwidth]{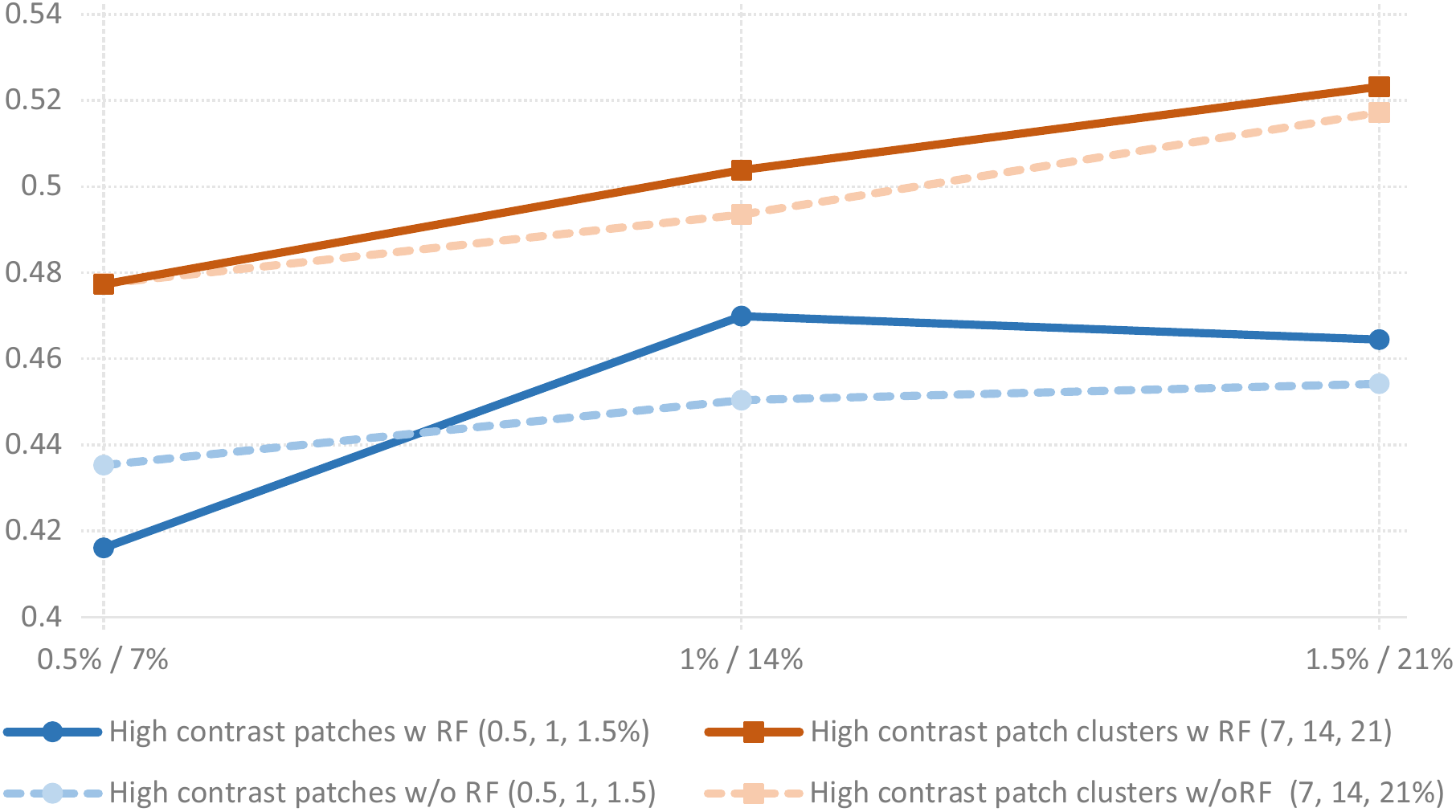}
    	\caption{Classification accuracy on the validation set at building-level for the different configurations of Table \ref{tab:sensitivity}.}
    \label{fig:val_accuracy}
\end{figure}

Age prediction aggregates the individual patch-wise age estimates over an entire building. Results for our approach from Section \ref{subsec:agePrediction} with majority voting as aggregation function for the validation set $T_v$ are shown in Figure \ref{fig:val_accuracy}. Similarly to the results in Table \ref{tab:sensitivity} more patches (higher percentages) lead to better performance and relevance filtering (solid lines) in most cases improves performance. The clustering-based patch selection clearly outperforms the purely contrast-based selection (orange line with square markers). The obtained accuracy on the validation set of 52\% clearly outperforms the random baseline of approx. 18\% (according to zero rule) as well as the patch-level accuracy of approx. 40\% from Section \ref{sec:preliminary}. These results show that fusion of the patch-wise predictions for the entire image strongly contributes to the classification performance.

We further evaluate the two different aggregation functions as well as the filtering of ambiguous patches from Section \ref{subsec:agePrediction}. Majority voting slightly outperforms averaging the likelihoods by a small margin of 1-2\% accuracy which indicates that this strategy is more robust. The removal of ambiguous patches improves results slightly (around 1\%) but not consistently in all experiments. Peak performance is, however, obtained by filtering ambiguous patches.

In a next step, we compare different network architectures and the effect of their combination in an ensemble. Additionally to the ResNet architecture, we train Alexnet with the same inputs. The peak performance of Alexnet on the validation set is an accuracy of 47.5\%. Alexnet does not reach the performance level of ResNet (51.5\%). When both nets are combined in an ensemble, however, the overall performance on the validation set increases to 54.0\%. This represents the best result  obtained during training. Thus, this configuration is used in all further experiments on the test sets.

\subsection{Generalization Ability and Dataset Bias}
\label{subec:genealization}
 
\begin{figure}[ht]
	\centering
    \subfigure[Test set $T_t$]{
		\includegraphics[width=0.47\linewidth]{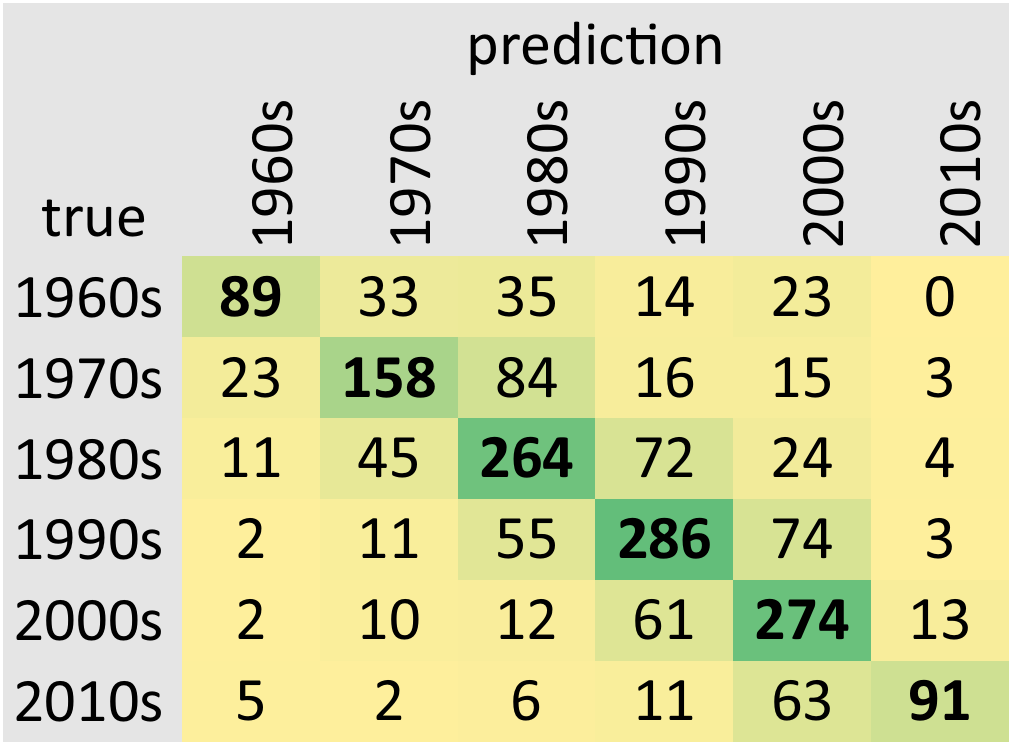}
    	\label{sfig:CMs}
	}
    \subfigure[Test set $W_t$]{
		\includegraphics[width=0.47\linewidth]{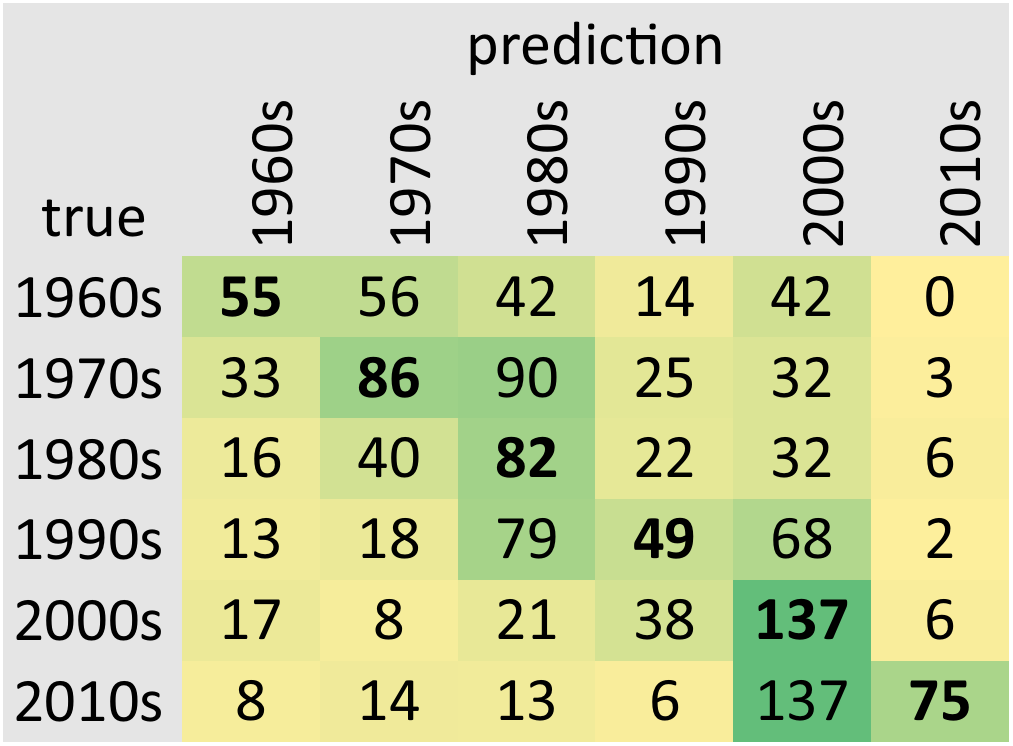}
    	\label{sfig:CMw}
	}
\caption{Confusion matrix for age prediction on the two employed test sets: (a) the approach generalizes well to test set $T_t$ with an overall accuracy of 61.35\%; (b) the additional complexity of dataset $W_t$ (inclusion of renovated buildings) represents an notable challenge and results in frequent underestimations of building age.}
\label{fig:confMatrix}
\end{figure}

After having evaluated the different components of our approach we evaluate it on the independent test set $T_t$ to estimate its generalization ability. We achieve an overall accuracy for building age prediction of 61.35\%. This clearly outperforms the best result obtained for the validation set (54\%, see Section \ref{subsec:agePredictionResults}). This proves a good generalization ability of our approach. A confusion matrix for the experiment is shown in Figure \ref{sfig:CMs}. The highest values are obtained along the diagonal (correct predictions). Most confusions occur between neighboring classes, which is due to the fuzziness at the epoch boundaries (decades).

Additionally, we test our approach on the second test set $W_t$, which comes from another data source (web images, see Section \ref{subsec:dataset}) to estimate the influence of dataset bias and obtain an accuracy of 34.94\%. This performance is notable smaller and shows that there is an influence from the data source \cite{torralba2011unbiased}. We assume the main reason for the drop in performance is that the test set $W_t$ contains also renovated and modernized buildings (see Section \ref{subsec:dataset}). Renovations change the visual appearance
of a building and usually make it look newer. This is reflected by the confusion matrix in Figure \ref{sfig:CMw}, which is skewed towards age predictions which underestimate the age (confusions in the upper triangular part of the matrix). This is particularly noticeable for older buildings (1960s-1990s). 

\subsection{Human Baseline Comparison}
\label{subsec:HumanBaselineresults}
To better understand the complexity of the task of visual building age estimation we set up an experiment utilizing LabelMe \cite{russell2008labelme}
to generate a human baseline, see also Section \ref{subsec:humanBaselineSetup}. The experiment has been performed within the scope of a case study at the Real Estate Industry Department of the University of Applied Sciences in Kufstein, Austria 
during winter term 2017 with a group of seven bachelor students from the Facility  \& Real Estate Management curriculum. 
The dataset $H_t$ was split into seven partitions. Each partition was assigned to one evaluator who assessed the building epoch for each house by looking at the respective photo and selecting an epoch from a list. 

For a fair and objective comparison, the same data was then provided to our pre-trained method. The human evaluators achieved an overall accuracy of 36\% (170 correct assignments from 470). This result shows that building age assessment is far from being trivial and poses a considerable challenge to humans. Our approach achieves an accuracy of 55.1\% on the same data (without any adaption or re-training) and thus strongly outperforms human performance. This demonstrates strong capabilities of our approach for automated age estimation. It further confirms a good generalization ability on $H_t$ which is an independent test set similarly to $T_t$.

\subsection{Network Analysis}
\label{sec:qualitativeResults}

\begin{figure}
	\centering
    \subfigure[most likely]{
		\includegraphics[height=0.34\linewidth]{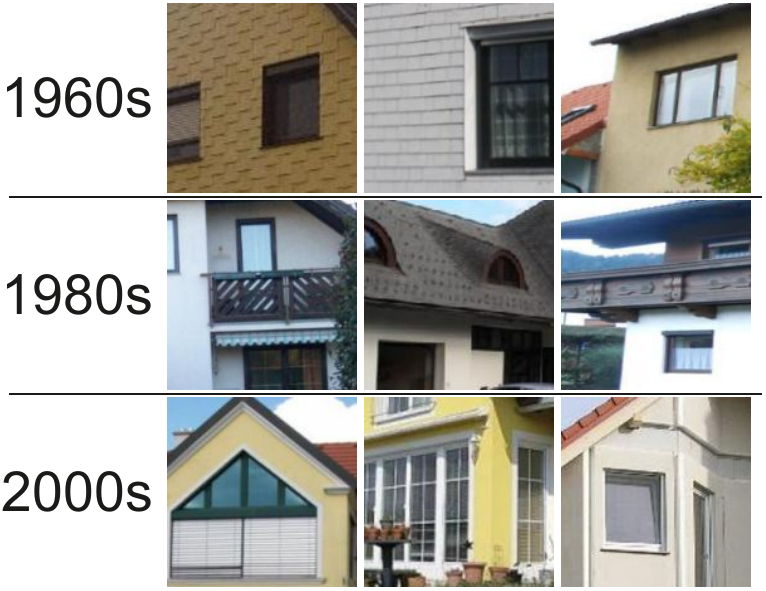}
    	\label{sfig:patches_top}
		}
    \subfigure[most unsure]{
		\includegraphics[height=0.34\linewidth]{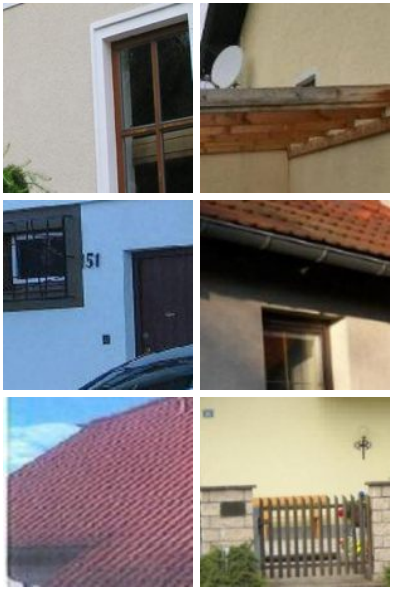}
    	\label{sfig:patches_unclear}
		}
    \subfigure[noise patches]{
		\includegraphics[height=0.34\linewidth]{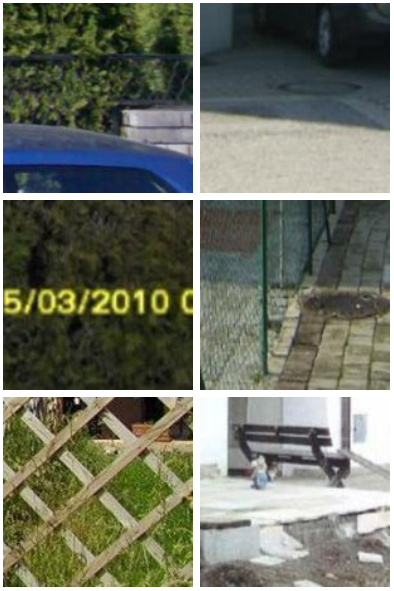}
    	\label{sfig:patches_wrong}
		}
    \caption [Caption for LOF]{Test patches from our experiment: (a) patches classified correctly by our network (ResNet) with highest likelihood for the correct class; (b) patches with lowest confidence for a certain class; (c) non-building patches that pass the relevance filtering.}
    \label{fig:patches_results}
\end{figure}

To get a deeper understanding on the inner workings of the trained networks and to verify the learning process, we perform a series of qualitative analyses. In a first step, we investigate the capabilities of the network to learn discriminative patches for certain building epochs (\textbf{RQ3} from Section \ref{subsec:RQs}). To this end, we extract the top 100 patches from our test set $T_t$ that were assigned to the correct class with highest probability, i.e. those patches the network is most confident about. Figure \ref{sfig:patches_top} shows example patches from three selected building epochs (1960s, 1980s, and 2000s) which achieve high likelihoods for their respective class ($\geq 0.99$). We observe, that a majority of the extracted patches represent typical building characteristics which confirms that the network is able to learn meaningful visual primitives. The experiment further shows that the most characteristic patches capture a rather large scale and thus larger parts of the buildings. This indicates that a larger spatial context facilitates capturing discriminative patterns. 

Figure \ref{sfig:patches_unclear} shows those patches the network is most uncertain about. From visual inspection we conclude that there are two major reasons for the uncertainty. First, the network seems to be unable to clearly decide for a class when there is a lack of structure as e.g. in the case of the patch which contains mostly the roof in the bottom-left corner of Figure \ref{sfig:patches_unclear} and second when the scale is to small and the spatial context is insufficient. Larger scales seem to facilitate finding typical age-related patterns. Finally, Figure \ref{sfig:patches_wrong} shows patches which unintendedly pass the relevance filtering. A common pattern observed for such false positives patches is a strong contrast across the patch and a rich structure in the image (e.g. the patch in the bottom left corner of Figure \ref{sfig:patches_wrong}.

\begin{figure}[ht]
\includegraphics[width=0.9\linewidth]{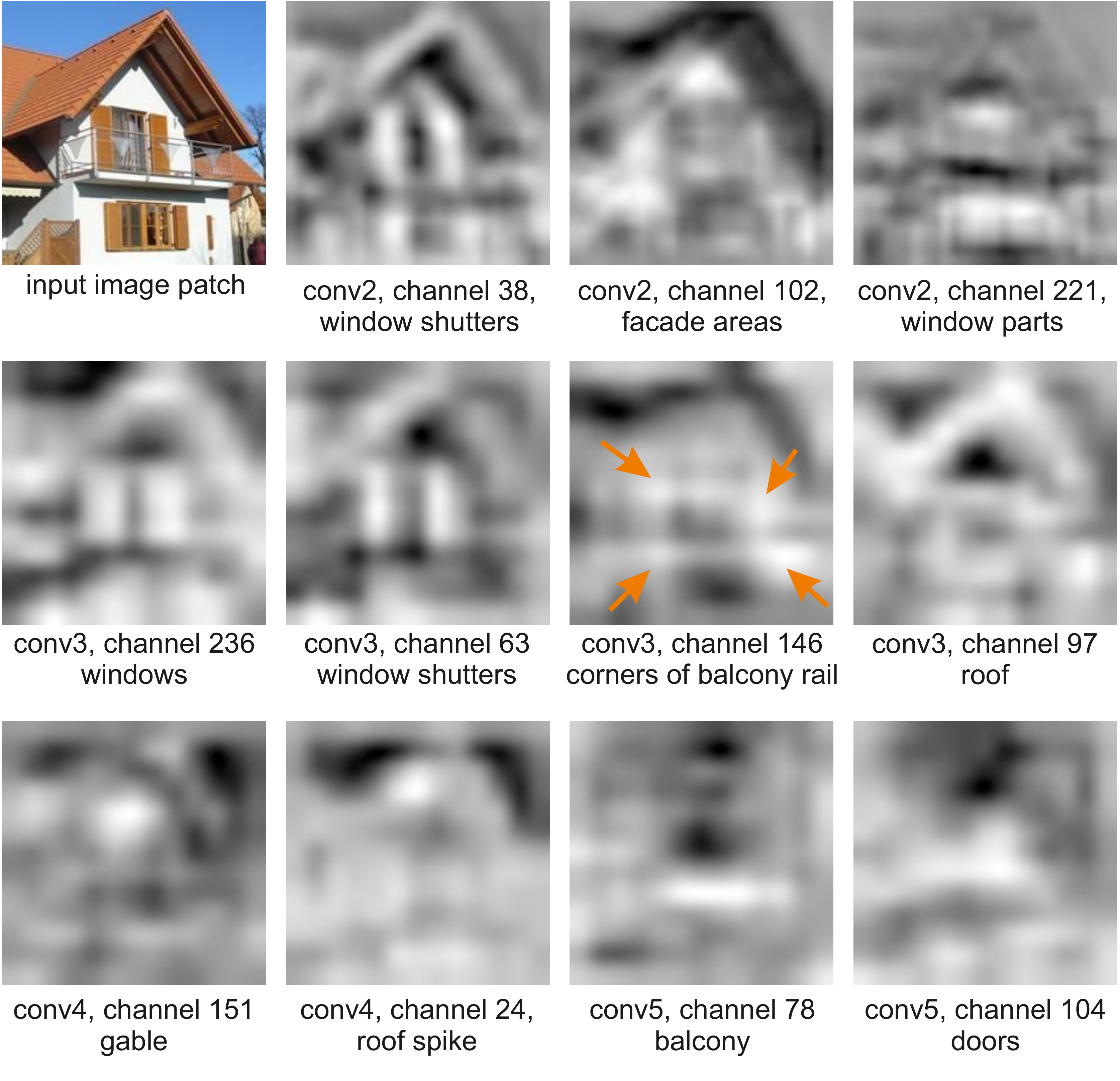}
\caption{Activations for an input patch of different convolutional layers of our trained network (2nd to 5th convolutional layer)  capturing different architectural elements of buildings.}
\label{fig:activations}
\end{figure}

We further investigate to which degree the learned filters in our networks reflect patterns that can be assigned to particular elements of buildings. To this end, we input patches to the network, perform a forward pass and extract  the activations for all convolutional layers. For visualization we scale them up to the input image size. Results of selected filters for an example patch are shown in Figure \ref{fig:activations}. We observe that building-related patterns are captured already in the second convolutional layer. In higher layers more specific structures are represented such as the spike of the roof and the gable (the triangular part of the wall underneath the roof) in the 4th convolutional layer, see Figure \ref{fig:activations}, first two images in last row.

\begin{figure}[ht]
\includegraphics[width=0.8\linewidth]{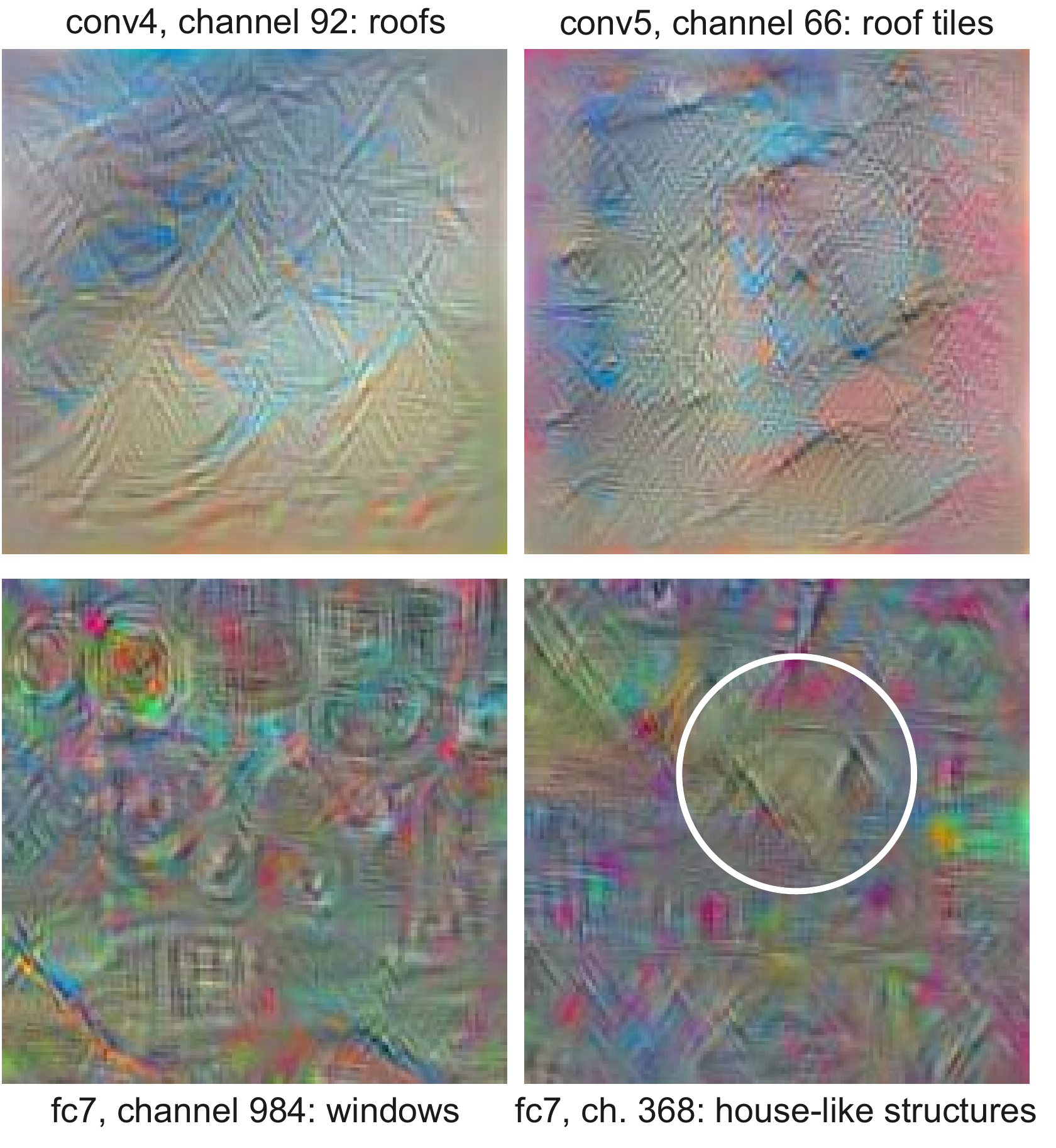}
\caption{Filters reconstructed with DeepDream capturing building-related patterns, such as windows, roofs and typical building contours (best viewed in color).
}
\label{fig:deepDream}
\end{figure}

To verify which structures are actually learned by the individual filters, we apply the DeepDream approach \cite{deepDream2015} to the individual convolutional filters. We apply DeepDream with three pyramid levels with a spacing of 1.05 and run the back propagation for 20 iterations. Results for representative filters learned are shown in Figure \ref{fig:deepDream} (best viewed in color). We frequently observe triangular roof-like structures, as in the top-left image in Figure \ref{fig:deepDream}. Other filters represent textures, such as areas covered by roof tiles (top-right image in Figure \ref{fig:deepDream}). In higher-layer filters, we observe larger structures, such as windows (bottom-left image) but even house-like structures (see white circle in bottom-left image in Figure \ref{fig:deepDream}).

\section{Conclusion}

Real estate image analysis (REIA) is a research field that attracts increasing attention. In this context, we have presented a first method for the estimation of building age  from unconstrained photographs. We mine promising image patches from building images at different scales and learn representations for different building epochs using CNNs. Our results show that different epochs can be discriminated with a probability of approx. 61\% which is far beyond random probability (18\% according to zero rule). Our approach is able to capture meaningful building characteristics for different building epochs. Furthermore, we clearly beat the human baseline and set a first automatic analysis baseline for this task. A major insight from the in-depth analysis of the trained networks is that the discovered structures represent more abstract patterns than those reported by e.g. object recognition networks \cite{deepDream2015}. The reason for this is that the classes employed in this work (building epochs) have a much higher intra-class variation than typical object classes (e.g. cats, cars, and dogs). The presented task is thus more difficult because it requires not only to learn a particular type of object (e.g. house) but also its typical characteristics that differentiate it from other objects of the \textit{same} class (e.g. house from the 1960s vs. house from the 1970s). To further improve results we will investigate deeper network architectures to obtain more powerful hierarchies, the effect of a more fine-grained ground truth for training (i.e. annotations of individual building elements instead of coarse image-level categories), and building segmentation techniques to better filter out non-building patches. Furthermore, we will investigate age prediction for renovated houses in detail to see whether the date of renovation can be predicted visually and to which degree indications for the original YoC can be inferred for such buildings.

\section{Acknowledgments}
We thank Sprengnetter Austria GmbH for providing real estate images and meta-data for our experiments. This work was supported by the Austrian Research Promotion Agency (FFG), Project No. 855784 and Project No. 856333.

\bibliographystyle{ACM-Reference-Format}
\bibliography{sample}

\end{document}